\documentclass[preprint,12pt]{elsarticle}

\usepackage{times}
\usepackage{epsfig}
\usepackage{graphicx}
\usepackage{amsmath}
\usepackage{amssymb}
\usepackage{multirow}
\usepackage[english]{babel}
\usepackage {subcaption}
\usepackage{float}

\def\etal{\emph{et al}.}
\def\eg{\emph{e.g}.}

\journal{arxiv.org}
\begin{document}

\begin{frontmatter}

\title{Adaptive 3D Face Reconstruction from a Single Image}
\author[1]{Kun Li}
\author[1]{Jing Yang}
\author[1]{Nianhong Jiao}
\author[1]{Jinsong Zhang}
\author[2]{Yu-Kun Lai}

% note need leading \protect in front of \\ to get a newline within \thanks as
% \\ is fragile and will error, could use \hfil\break instead.
%E-mail: lik@tju.edu.cn
\address[1]{Tianjin University,
Tianjin 300350, China.}

\address[2]{Cardiff University, Cardiff CF24 3AA, UK.}

% for Computer Society papers, we must declare the abstract and index terms
% PRIOR to the title within the \IEEEtitleabstractindextext IEEEtran
% command as these need to go into the title area created by \maketitle.
% As a general rule, do not put math, special symbols or citations
% in the abstract or keywords.

\begin{abstract}
3D face reconstruction from a single image is a challenging problem, especially under partial occlusions and extreme poses. This is because the uncertainty of the estimated 2D landmarks will affect the quality of face reconstruction. In this paper, we propose a novel joint 2D and 3D optimization method to adaptively reconstruct 3D face shapes from a single image, which combines the depths of 3D landmarks to solve the uncertain detections of invisible landmarks. The strategy of our method involves two aspects: a coarse-to-fine pose estimation using both 2D and 3D landmarks, and an adaptive 2D and 3D re-weighting based on the refined pose parameter to recover accurate 3D faces. Experimental results on multiple datasets demonstrate that our method can generate high-quality reconstruction from a single color image and is robust for self-occlusion and large poses.
\end{abstract}

% Note that keywords are not normally used for peerreview papers.

\end{frontmatter}

\section{Introduction}
\label{sec:introduction}

Human reconstruction from images, especially for faces, is an important and challenging problem, which has drawn much attention from both academia and industry \cite{fang2020face, liang2017pose, li2013data}. Although existing face reconstruction methods based on multiple images have achieved promising results, it is still a tough problem for a single input image, especially under partial occlusions and extreme poses.

3D Morphable Model (3DMM) ~\cite{blanz1999morphable, egger20193d} is a popular and simple linear parametric face model. Some methods~\cite{Luan2018Nonlinear, Tran2016Regressing, Jourabloo2016Large, amberg2008expression} achieve 3D face reconstruction from a single image using convolutional neural networks (CNN). To fit 3DMM to a facial image with self-occlusions or large poses, Zhu \etal \cite{Zhu2016Face} and Yi \etal \cite{yi2019mmface} take a 3D solution to reconstruct the face with 3D landmarks. However, these methods ignore the effect of 2D landmarks for visible parts which are more accurate. Moreover, lack of enough 3D face datasets with ground-truth for training limits the performance of these learning-based methods. By contrast, traditional optimization-based methods~\cite{Chen2014Displaced, Saito2016Real, Garrido2016Reconstruction, Luo20183D} are more flexible to fit the 3DMM model. But these methods heavily depend on accurate 2D landmark detection, and tend to generate poor or incorrect face reconstruction for facial images with occlusions. To address the occlusion problem, Lee \etal \cite{Lee2012Single} and Qu \etal \cite{Qu2014Fast} discard the occluded landmarks, but their methods lack constraints of complete landmarks. To fix 2D landmark correspondence errors caused by face orientation or hair occlusion, Zhu \etal \cite{Zhu2015High} and Luo \etal \cite{Luo20183D} propose landmark marching methods to update silhouette vertices. However, they need to manually label 68 landmark vertices, which is laborious and time consuming. Due to the lack of depth information, these traditional methods are still hard to correctly reconstruct invisible areas, and hence difficult to deal with extreme poses, \eg, $90^\circ$ side faces.

\begin{figure}[t]
  \centering
		\includegraphics[width=1.0\linewidth]{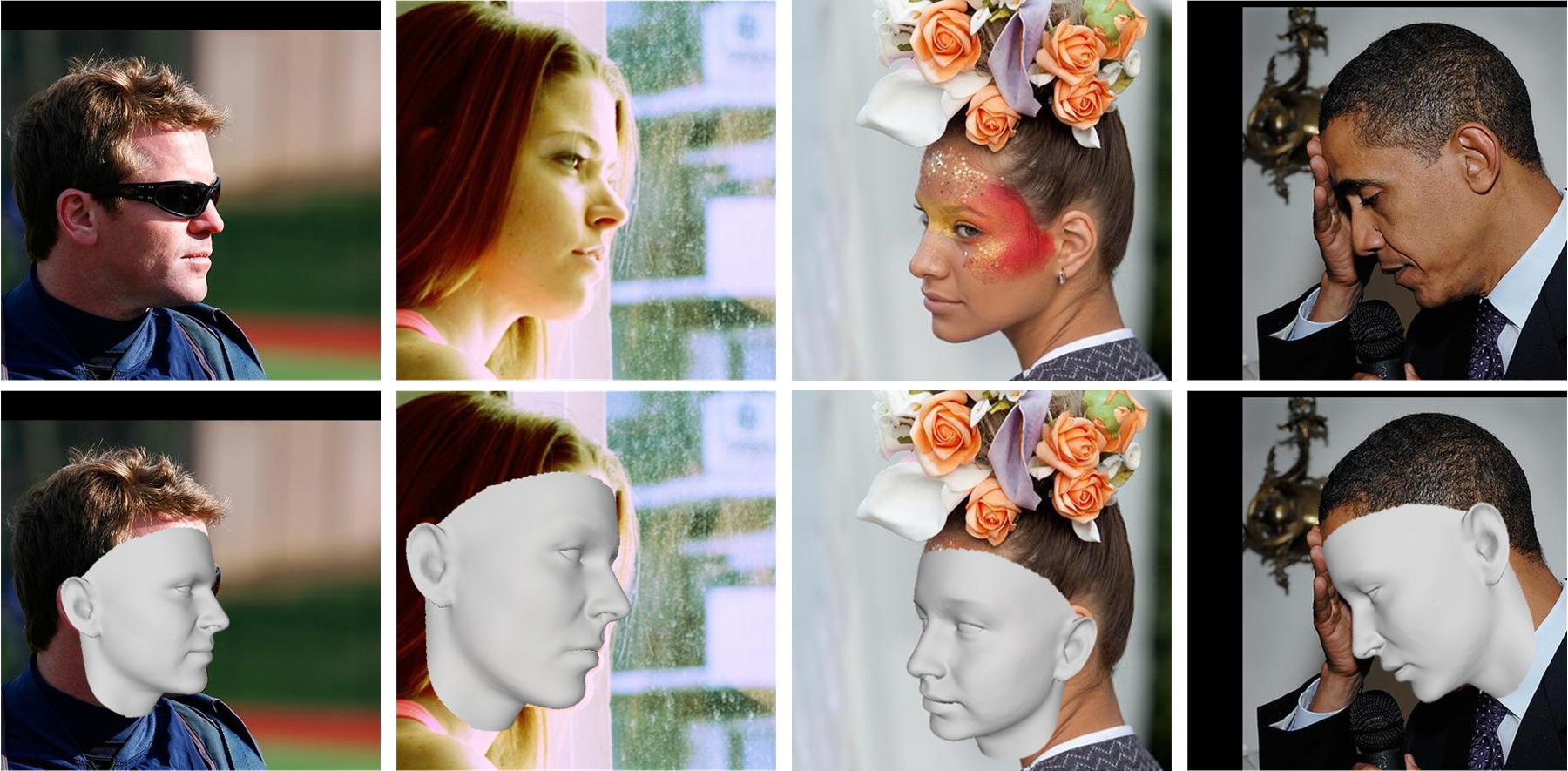}
\caption{3D face reconstruction results from single images using our method. Our method is robust to extreme poses and partial occlusions.}
\label{fig:result}
\end{figure}

\begin{figure*}[tbp]
  \centering
  \includegraphics[width=1\linewidth]{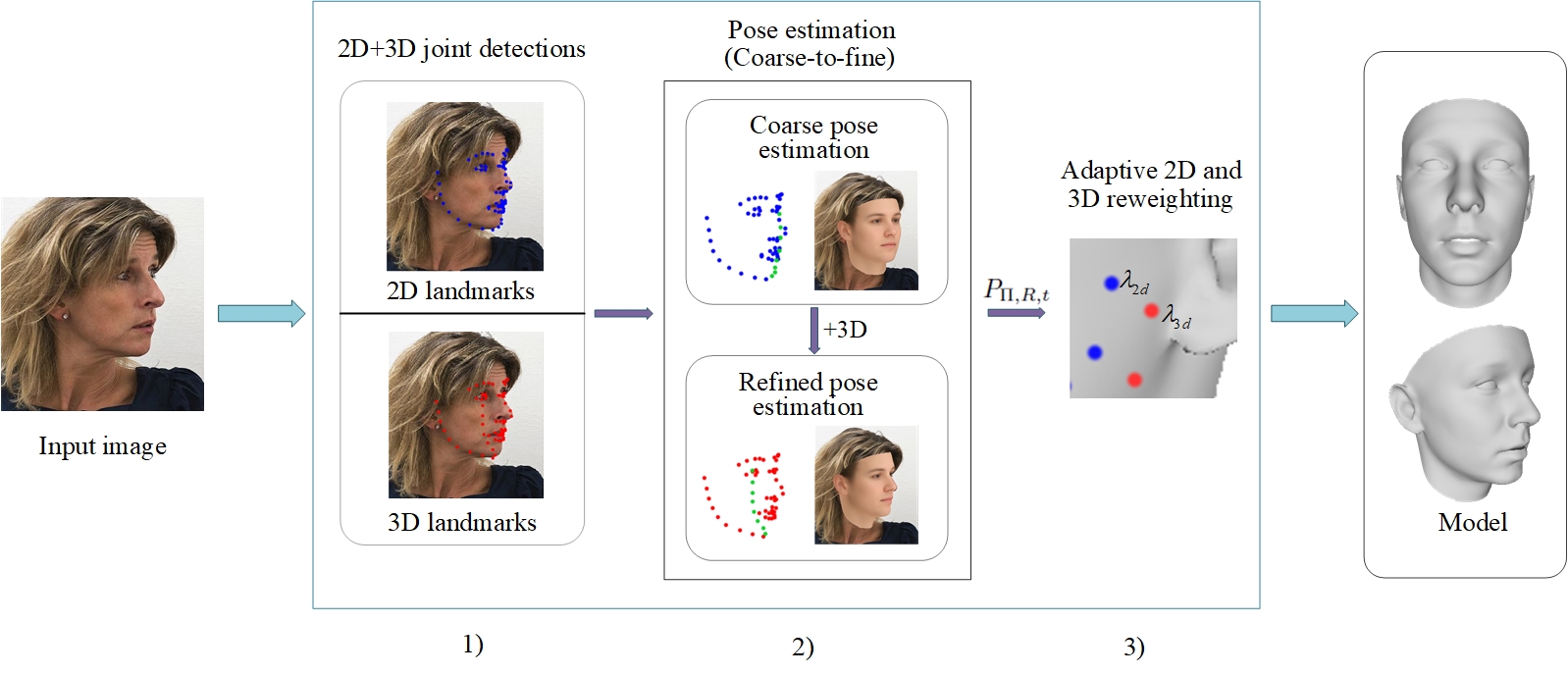}
  \caption{ {The pipeline of our method: 1) 2D and 3D landmark detection; 2) Coarse-to-fine pose estimation: Coarse pose estimation includes 68 2D landmarks (the occluded silhouette landmarks are shown in green and the estimated coarse pose with mean face is shown in the input image), and the pose is refined by combining them with the estimated 3D silhouette landmarks (the occluded 3D silhouette landmarks are shown in green and the estimated refined pose with mean face is shown in the input image); 3) Adaptive 2D and 3D re-weighting: $\lambda _{2d}$ and $\lambda _{3d}$ are 2D and 3D weights, respectively. 2) and 3) are regarded as a bundle to achieve 3D face reconstruction. }}
  \label{fig:pipeline}
\end{figure*}

Inspired by recent work on 3D landmark detection, we use the depth information of 3D landmarks together with 2D landmarks to resolve the inherent depth ambiguities of the re-projection constraint during 3D face reconstruction by joint 2D and 3D optimization. 2D landmarks give the pixel positions of facial silhouettes based on the input image, while 3D landmarks give the depth positions of the facial silhouettes. It is hard to decide which detected landmarks are more believable. In order to effectively combine 2D and 3D landmarks, we propose a 2D and 3D re-weighting method to adaptively adjust the weights of 2D and 3D landmarks. In addition, instead of solving pose parameters directly, we design a coarse-to-fine method for accurate face pose estimation. Our method does not need manual intervention, and is robust to extreme poses and partial occlusions. Experimental results demonstrate that our method outperforms the state-of-the-art methods on AFLW2000~\cite{Zhu2016Face} and MICC~\cite{Bagdanov2012Florence} datasets, especially for non-frontal images. Figure~\ref{fig:result} shows some 3D face reconstruction results using our method.

Our main contributions are summarized as follows:
\begin{itemize}
\item \textbf{Joint 2D and 3D optimization}.
We formulate the 3D face reconstruction problem in a unified joint 2D and 3D optimization framework. To our best knowledge, our method is the first optimization method using both 2D and 3D information for face reconstruction. Our method is fully automatic and robust to extreme poses and partial occlusions.

\item \textbf{Coarse-to-fine pose estimation}.
To obtain accurate pose parameters for face reconstruction, we propose a coarse-to-fine scheme using both 2D and 3D landmarks. We generate a coarse pose estimation by fitting the 3DMM model with the silhouettes of 2D landmarks and obtain a refined pose estimation by replacing the invisible 2D landmarks with the corresponding 3D silhouettes.

\item \textbf{Adaptive 2D and 3D re-weighting}.
We propose an adaptive 2D and 3D re-weighting scheme to adaptively adjust the weights of 2D and 3D landmarks according to the acquired pose estimation. Among them, 2D landmarks are sufficiently accurate for visible areas, and the depth information of 3D landmarks will improve the detection accuracy for invisible areas. For example, the weights of 2D landmarks should be increased under small poses while the weights of 3D landmarks should be increased under large poses. To achieve this, we provide two adaptive weight adjustment schemes to deal with small-pose and large-pose, respectively.
\end{itemize}

\section{Related Work}
\label{sec:related work}

Over the years, many methods solve the face reconstruction problem caused by self-occlusions or head rotations with multiple images. Although these methods achieve promising results, the requirement of multiple inputs limits their practical applications. It is more prospective and challenging to reconstruct a 3D face from a single image.
In this section, we review the related work on 3D face reconstruction from a single image.

\subsection{2D and 3D Face Alignment}

Most of early face alignment methods can only roughly detect the 2D face landmarks, until the emergence of new techniques based on cascaded regression \cite{Cao2014Face, Xiong2013Supervised, Zhu2016Face,  Bulat2016Convolutional}. This kind of methods largely improves the accuracy of 2D face alignment methods and performs well on the LFPW~\cite{Belhumeur2013Localizing} and 300-W~\cite{Sagonas2013semi} datasets. With the development of convolutional neural networks (CNNs), Sun \etal \cite{Sun2013deep} propose to acquire 68 facial landmarks by a CNN cascade method. Multi-task learning and attribute classification are combined with CNN to obtain better results~\cite{Zhang2014Facial}. However, these methods are mainly effective for near-frontal faces. Some methods are proposed to solve 3D face alignment~\cite{Jourabloo2016Large, Zhu2016Face} that works better on large poses. Bouaziz \etal \cite{bouaziz2014dynamic} propose an algorithm  of 2D/3D registration based on RGB-D devices. Yi \etal \cite{Bulat2017How} take an image and 2D landmarks as inputs and use a 2D-to-3D network to learn the corresponding 3D landmarks, which can detect both the 2D landmarks and 3D landmarks.

\subsection{Single-view 3DMM-based Face Reconstruction}

3D Morphable Model (3DMM) is first proposed by Blanz and Vetter \cite{blanz1999morphable}, and improved to have expression parameters by using 3D FaceWarehouse~\cite{Chen2014FaceWarehouse}. 3DMM has a wide range of applications due to its flexibility and convenience by adjusting parameters to present different face shapes and expressions. Given a single color image, optimization-based methods~\cite{Roth2015Unconstrained, Huber2016A,  Lee2012Single, Luo20183D} estimate the 3DMM face by constraining the data similarities of facial landmarks, lighting or edges. Recently, learning-based approaches~\cite{Thies2016Demo, Jackson2017large} have been proposed to deal with the single-image reconstruction problem. Tran \etal \cite{Tran2016Regressing} propose a regression-based method to refine the 3DMM parameters, and Kim \etal \cite{Kim_2018_CVPR} design deeper networks to obtain more discriminative results. Yi \etal \cite{yi2019mmface} propose an end-to-end method including a volumetric sub-network and a parametric sub-network to reconstruct a face model, which separates the identity and expression parameters. However, lack of enough 3D face datasets with ground-truth for training limits the performance of these learning-based methods.

\subsection{Landmark Updating Method}

To fix the landmark fitting error caused by large poses and self-occlusions, Lee~\cite{Lee2012Single} and Qu \etal \cite{Qu2014Fast} propose to discard invisible landmarks, but these methods cannot make full use of landmark constraints. Asthana \etal \cite{asthana2013robust} propose a look-up table containing 3D landmark configurations for each pose, but this method depends on pose estimation and need to build a large table in unconstrained environment. Zhu \etal \cite{Zhu2015High} first propose a landmark marching method which intends to move the 3D landmarks along the surface to rebuild the correspondence of 2D silhouette automatically. Zhang \etal \cite{Luo20183D} update the silhouette landmark vertices by constructing a set of horizontal lines and choosing among them a set of vertices to represent the updated silhouette. A common disadvantage of their approaches is that they need to manually label many landmarks, which takes a lot of time and effort. Moreover, when the deflection angle becomes larger, the detected 2D landmarks of the invisible face part will be less accurate. Even if the silhouette is updated, the methods do not work well for large-pose images with a deflection angle larger than $60^\circ$.

In this paper, we propose a novel automatic 3DMM-based face reconstruction method from a single image by joint 2D and 3D optimization, which is robust to extreme poses and self-occlusions.

\section{Method}
\label{sec:model}
It is difficult to accurately detect 2D landmarks along the face silhouette under large poses or partial occlusions, but 3D depth information of the face provides strong constraints even for invisible landmarks. Therefore, our method solves the face reconstruction problem under large poses or partial occlusions by joint 2D and 3D optimization. Figure~\ref{fig:pipeline} illustrates the pipeline of our method. Our method reconstructs a 3D face model from a single image based on 3DMM~\cite{blanz1999morphable}. For an input image, we first detect the 2D and 3D positions of the 68 landmarks using an efficient detection method~\cite{Bulat2017How} that provides both 2D and 3D landmarks for the same image.

Traditional pose estimation methods are difficult to accurately obtain the face pose due to the errors in detection of 2D landmarks in the occluded regions. We propose a coarse-to-fine pose estimation scheme using both 2D and 3D landmarks. In the coarse step, we estimate Euler angles using the left and right silhouette landmarks respectively, and then choose the maximum value as the initial pose. In the refined step, 68 landmarks are updated by replacing the 2D landmarks of the invisible silhouette landmarks with the corresponding 3D landmarks. In order to make full use of 3D depth information and 2D position information, we propose to automatically adjust 2D and 3D weights with an adaptive re-weighting scheme. We regard the pose estimation and adaptive re-weighting as a bundle to reconstruct the 3D face geometry.

\subsection{3D Morphable Model}\label{sec:3dmm}

3D Morphable Model is a 3D face statistical model, which is proposed to solve the problem of 3D face reconstruction from 2D images. In this work, we merge the Basel Face Model (BFM)~\cite{paysan20093d} and the Face Warehouse~\cite{Cao2014FaceWarehouse} with non-rigid ICP~\cite{amberg2007optimal} to construct our 3DMM. It is a linear model based on Principal Components Analysis (PCA) which describes the 3D face space as
\begin{eqnarray}\label{1}
M=\bar{m}+\Gamma_{sha} \alpha+\Gamma_{\exp } \beta,
\end{eqnarray}
where $M$ represents a 3D face, $\bar{m}$ is the mean shape, $\Gamma _{sha}$ is the principal axes corresponding to face shapes coming from BFM~\cite{paysan20093d} and $\alpha$ is the shape parameter. $\Gamma _{\exp }$ is the principal axes corresponding to face expressions coming from Face Warehouse~\cite{Cao2014FaceWarehouse} and $\beta$ is the expression parameter. The collection of pose parameters is $P_{\Pi, R, t}$, where $R$ is a $3 \times 3$ rotation matrix constructed from rotation angles ($pitch$, $yaw$, $roll$) and $t$ is a $3\times 1$ translation vector. The projection matrix $\Pi$ is formulated as
\begin{eqnarray}\label{2}
\Pi  = s\left[ {\begin{array}{*{20}{c}}
1&0&0\\
0&1&0
\end{array}} \right],
\end{eqnarray}
where $s$ is the scale factor.
The 2D projection of the 3D face model with weak perspective projection ~\cite{Bruckstein1999Optimum} is represented as
\begin{eqnarray}\label{3}
l_{2 d}(\alpha, \beta)=\Pi R\left(\bar{m}+\Gamma_{s h a} \alpha+\Gamma_{\exp } \beta\right)+t.
\end{eqnarray}

\subsection{Joint 2D and 3D Optimization}
\label{sec:joint_optimization}

Traditional face reconstruction methods depend on the detected 2D landmarks which have low accuracy for non-frontal images, especially for very large poses. In order to solve the problem of inaccurate detection of 2D silhouette landmarks in non-frontal images, we propose a joint 2D and 3D optimization method. Specifically, we propose a coarse-to-fine pose estimation method using both 2D and 3D landmarks, and then iteratively optimize them with the projected 3D vertices. In order to handle various rotation angles, we propose an adaptive reweighting method.

We solve the fitting process by joint 2D and 3D optimization and take the shape and expression prior terms into a hybrid objective function. We formulate it as a nonlinear least squares problem:
\begin{eqnarray}\label{4}
{{E}_{fit}}\left( \alpha ,\beta ,P_{\Pi,R,t} \right)={{\lambda }_{2d}}{{E}_{2d}}\left( \alpha ,\beta ,P_{\Pi,R,t} \right)\nonumber \\
+{{\lambda }_{3d}}{{E}_{3d}}\left( \alpha ,\beta ,P_{\Pi,R,t} \right)
+{{E}_{p}}\left( \alpha ,\beta \right),
\end{eqnarray}
where $\lambda _{2d}$ and $\lambda _{3d}$ are 2D and 3D weights, respectively. ${{E}_{2d}}\left(\alpha, \beta, P_{\Pi,R,t}\right)$ and ${{E}_{3d}}\left(\alpha, \beta ,P_{\Pi,R,t} \right)$ are the alignment energies based on regressed 2D and 3D landmarks, respectively, which will be elaborated in Section~\ref{sec:fitting_reweighting}. ${{E}_{p}}\left( \alpha ,\beta  \right)$ is a prior term of both shape and expression, which will be explained in Section~\ref{sec:shexp_priors}.

We first initialize the shape parameter $\alpha$ and expression parameter $\beta$ with zeros, and then use our coarse-to-fine pose estimation method to estimate a coarse pose and a refined pose. Based on the pose estimation, we finally solve the optimization problem to obtain shape and expression parameters iteratively. After each iteration, After each iteration, we get a new model with the updated shape and expression parameters, and then re-estimate pose parameters. This process iterates four times in our experiments, which is sufficient to converge in practice.

\subsubsection{Coarse Pose Estimation}
We estimate a coarse pose $P_{c}$ by computing Euler angles using the 2D landmarks of the left or right silhouettes and the rest 51 landmarks respectively. There are 17 detected 2D face silhouette landmarks, which are divided into three parts according to the locations: left (1 to 9), middle (10), and right (11 to 17). If the Euler angle calculated by the left silhouette $P_{l}$ is greater than that by the right silhouette $P_{r}$, it means that the head orientation is to the left, and vice versa. We regard the max value $P_{c}=P_{l}$ or $P_{c}=P_{r}$ as the final face pose direction on the $Y$ axis.

\subsubsection{Refined Pose Estimation}\label{sec:refine_pose}
To resolve the inevitable depth ambiguities of 2D re-projection constraint, we add 3D constraint to improve the accuracy of pose estimation.
First, we project the mean face model onto the image plane to accurately capture the invisible 2D silhouette landmarks. Then, we replace the invisible 2D silhouette landmarks with the corresponding estimated 3D landmarks by the method~\cite{Bulat2017How}. Finally, we update the 68 landmarks and fit the parametric face model with our input image to get the refined pose parameters $P_{r e f}$. We regard the refined pose estimation as the initial value for the optimization, and in the next iteration, we will update pose parameters as $P_{\Pi, R, t}$.

\subsubsection{2D and 3D Fitting with Adaptive Re-weighting}\label{sec:fitting_reweighting}
For each input image, we detect the 2D and 3D landmarks $\{L_{2 d, i} \in \mathbb{R}^{2}\}_{1 \leq i \leq 68}$ and $\{L_{3 d, i} \in \mathbb{R}^{3}\}_{1 \leq i \leq 68}$ using an efficient detection method~\cite{Bulat2017How}.
The 2D fitting constraint ${E_{2d}}$ is defined as
\begin{eqnarray}\label{5}
E_{2 d}\left(\alpha, \beta, P_{\Pi, R, t}\right)=\sum_{i=1}^{68}\left\|l_{2 d, i}(\alpha, \beta)-L_{2 d, i}\right\|_{2}^{2},
\end{eqnarray}
where $l_{2 d, i}(\alpha, \beta)$ is the 2D projection coordinates of the $i$-th vertex of the 3D face model, as defined in Section \ref{sec:3dmm}. $L_{2d,i}$ is the $i$-th detected 2D landmark. We solve the 3DMM parameters by minimizing the Euclidean distances between the detected landmarks and the 2D projections of 3D points. We further incorporate the 3D depth information into the optimization to solve the ambiguities of invisible face area, by proposing a 3D alignment term as follows:
\begin{eqnarray}\label{6}
E_{3 d}\left(\alpha, \beta, P_{\Pi, R, t}\right)=\sum_{i=1}^{68}\left\|l_{3 d, i}(\alpha, \beta)-\left(L_{3 d, i}+t^{\prime}\right)\right\|_{2}^{2},
\end{eqnarray}
where $l_{3 d, i}(\alpha, \beta)$ is the 3D position of the $i$-th face landmark, and $L_{3 d, i}$ is the $i$-th detected 3D landmark. $t^{\prime}\in \mathbb{R}^{3}$ is an auxiliary variable that transforms the $L_{3 d, i}$ to the global coordinate system. The pose parameters and the optimization method are the same as $E_{2d}$. In order to effectively combine the 2D and 3D landmarks, we propose an adaptive weighting method:
\begin{eqnarray}\label{7}
W_{\lambda}=\left\{\begin{array}{ll}{1} & {\frac{2|yaw|}{\pi} \geq \varepsilon} \\ {0} & {\text { otherwise, }}\end{array}\right.
\end{eqnarray}
where $\varepsilon$ is set to 0.5, which means that we regard a $45^\circ$ angle as the boundary of head rotation for large pose and small pose.  ${W_\lambda=1}$ means large pose, and in that case, the 3D weight and the 2D weight are calculated as
\begin{eqnarray}\label{8}
{\lambda _{3d}} = \frac{{2|yaw|}}{\pi },
\end{eqnarray}
\begin{eqnarray}\label{9}
{\lambda _{2d}} = \left( {1 - \frac{{2|yaw|}}{\pi }} \right) \cdot w,
\end{eqnarray}
where $w$ is set to $0.5$. Otherwise, if ${W_\lambda}=0$, which means that the pose angle is less than $45^\circ$, the 3D weight and the 2D weight are calculated as
\begin{eqnarray}\label{10}
{\lambda _{3d}} = \frac{{2|yaw|}}{\pi } \cdot w,
\end{eqnarray}
\begin{eqnarray}\label{11}
{\lambda _{2d}} = 1 - \frac{{2|yaw|}}{\pi }.
\end{eqnarray}

\subsubsection{Shape and Expression Priors}\label{sec:shexp_priors}
We expect that each of the shape and expression parameters follows a normal distribution with zero mean and unit variance. The shape and expression prior terms are defined as
\begin{eqnarray}\label{12}
{E_p}\left( {\alpha ,\beta} \right) = {\lambda _\alpha }{E_{prior}}\left( \alpha  \right) + {\lambda _\beta }{E_{prior}}\left( \beta  \right),
\end{eqnarray}
where ${E_{prior}}\left( \alpha  \right)$ and ${E_{prior}}\left( \beta  \right)$ are shape and expression priors, respectively. ${\lambda _\alpha }$ and ${\lambda _\beta }$ are their corresponding weights. The shape prior is calculated as
\begin{eqnarray}\label{13}
E_{prior}\left(\alpha\right)=\left(\lambda_{2 d}+\lambda_{3 d}\right) \sum_{i=1}^{N_{\alpha}}\left(\frac{\alpha_{i}}{\sqrt{\delta_{\alpha_{i}}}}\right)^{2},
\end{eqnarray}
where ${N_\alpha }$ is the number of shape parameters, ${\alpha _i}$ is the $i$-th shape principal component, and ${\delta _{{\alpha _i}}}$ is the eigenvalue corresponding to the principal component. The expression prior is similarly defined as
\begin{eqnarray}\label{14}
E_{prior}\left(\beta \right)=\left(\lambda_{2 d}+\lambda_{3 d}\right) \sum_{i=1}^{N_{\beta}}\left(\frac{\beta_{i}}{\sqrt{\delta_{\beta_{i}}}}\right)^{2}.
\end{eqnarray}

Our joint 2D and 3D optimization algorithm is summarized in Algorithm 1, and we set $iters$ = $4$ in our experiments.

\begin{table}[!ht]
\label{alg}
\small
 \begin{center}
 \begin{tabular}{l}
  \hline
Algorithm 1: Joint 2D and 3D optimization. \\
 \hline
    1. \quad Input: a single image $I$.\\
    2. \quad Detect 2D and 3D landmarks;\\
    3. \quad Coarse pose estimation, $P_{c}$: \\
    4. \quad \quad Estimate $P_{l}$, $P_{r}$ based on left and right landmarks;\\
    5. \quad \quad Calculate Euler angle according to $P_{l}$, $P_{r}$ and get $yaw$,\\
    \quad \quad \quad $yl$: left $yaw$, $yr$: right $yaw$;\\
    6. \quad \quad If abs($yl$) $>$ abs($yr$), then\\
    7. \quad \quad \quad $P_{c}$ = $P_{l}$;\\
    8. \quad \quad Else\\
    9. \quad \quad \quad $P_{c}$ = $P_{r}$;\\
    10. \quad \quad End if\\
    11. \quad Refine pose estimation: \\
    12. \quad \quad Update 2D invisible silhouette landmarks and estimate\\
    \quad \quad \quad ~~the refined parameters, $P_{r e f}$; \\
    13. \quad $P_{\Pi, R, t}$ = $P_{r e f}$; \\
    14. \quad For $i$ = 1 to $iters$ do \\
    15. \quad \quad Update pose parameters, $P_{\Pi, R, t}$;\\
    16. \quad \quad Compute 2D and 3D weights adaptively;\\
    17. \quad \quad Estimate the expression parameter $\beta$ and shape\\ \quad \quad \quad ~~parameter $\alpha$;\\
    18. \quad End For\\
    19. \quad Output: $\{P_{\Pi, R, t}, M(\alpha,\beta)\}$\\
	\hline
 \end{tabular}
 \end{center}
\end{table}

\section{Experimental Results}
\label{sec:experiment}

In this section, we first introduce the datasets and metrics in Section \ref{sec:exp_results}, and demonstrate that our method is able to reconstruct 3D face from a single image under partial occlusions and various poses in Section \ref{sec:results}. Then, we perform an ablation study to analyze the effects of different components of our approach in Section \ref{sec:abl_study}. Finally, we compare our method with several state-of-the-art methods quantitatively and qualitatively in Section \ref{sec:compare}. More results can be found in the supplementary video.

\subsection{Datasets and Metrics}\label{sec:exp_results}

We conduct our qualitative experiments on AFLW2000~\cite{Zhu2016Face} dataset, which is a large-scale face database including multiple poses and perspectives. The MICC dataset~\cite{Bagdanov2012Florence} contains 53 videos with various resolutions, conditions and zoom levels for each subject. In order to demonstrate the effectiveness for partial occlusions and extreme poses, in ablation study, we select 106 non-frontal images with left or right view direction (53 images per case) in the Indoor-Cooperative videos for each subject to demonstrate the effectiveness of our each component quantitatively and qualitatively. For the comparison, we choose 53 the most frontal images and 53 non-frontal images for each subject from the videos as our test datasets. The reconstructed face is aligned with its corresponding ground-truth model using the Iterative Closest Point (ICP) method~\cite{Rusinkiewicz2001Efficient}, and we calculate the reconstruction errors for the face part by cropping the model at a radius of 85\emph{mm} around the tip of nose. 3D Root Mean Square Error (3DRMSE) is used to measure the model quality:
%%%YKL I think the original equation was wrong /N should be inside the square root. Please check the modified equation below. There are also many inconsistencies with the description below. Once you say it is the i-th face, but then i goes over all the vertices, as N is the number of vertices!
%%%YKL Please check the revised version below to make sure it is consistent with your implementation
\begin{eqnarray}\label{15}
\sqrt{\sum_{i}\left(\mathbf{X}-\mathbf{X}^{*}\right)^{2}} / N,
%\sqrt{\frac{1}{N}\sum_{i}\left(\mathbf{X}_{i}-\mathbf{X}^{*}_i\right)^{2}},
\end{eqnarray}
%%%modified
where $\mathbf{X}$ is the reconstructed face (after cropping), $\mathbf{X}^{*}$ is the ground truth, and $N$ is the number of vertices of the 3D model cropped from the reconstructed face.

\begin{figure*}[!t]
  \centering
  \includegraphics[width=0.9\linewidth]{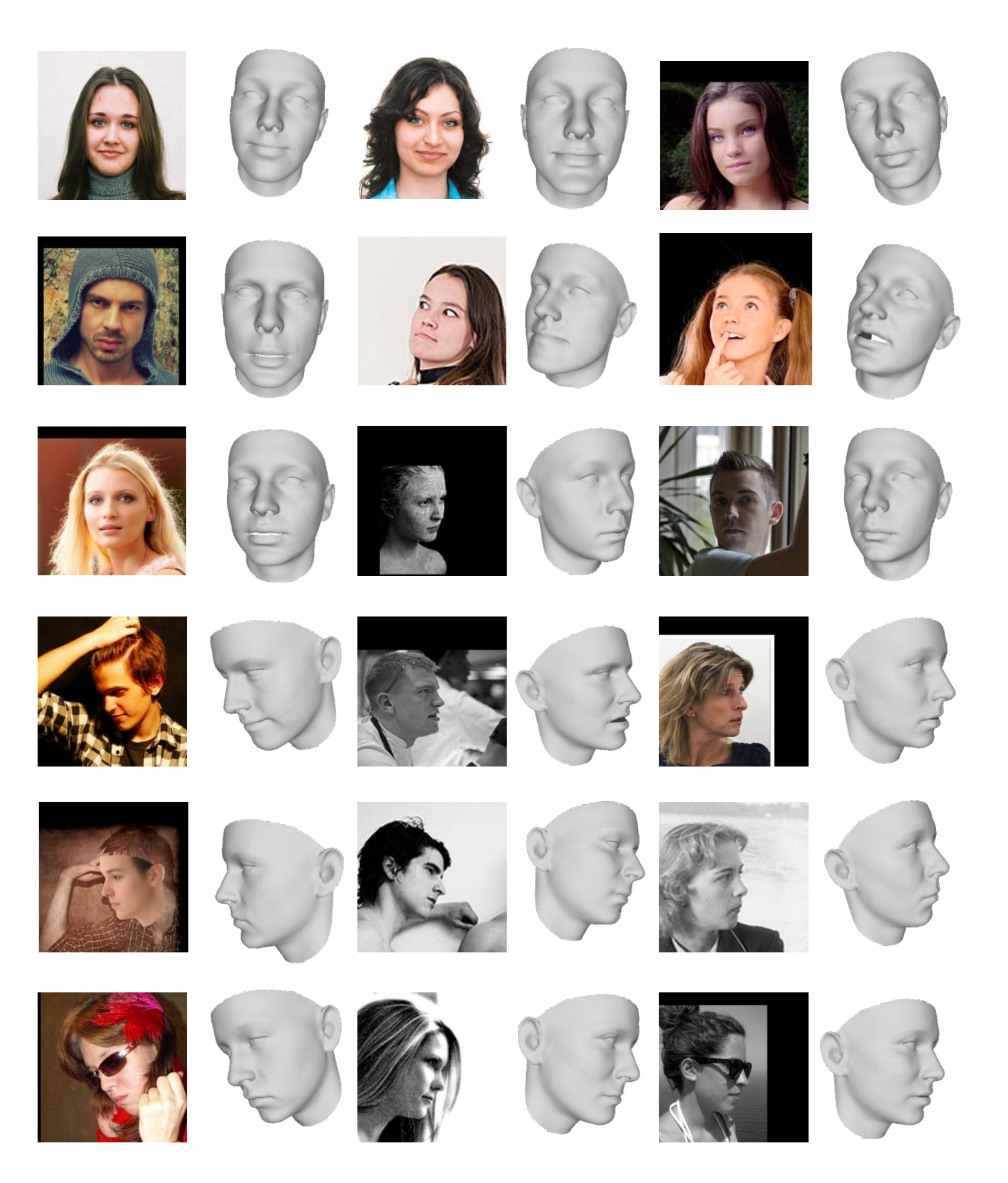}
  \caption{Qualitative results on the AFLW2000~\cite{Zhu2016Face} dataset. }
  \label{fig:bc3}
\end{figure*}

% \begin{figure*}[t]
% \begin{minipage}[b]{1.0\linewidth}
%   \centering
%   \centerline{\epsfig{figure=OC1.jpg,width=8.5cm}}
%   % \vspace{1.5cm}
%   % \centerline{(b) Results 2}\medskip
% \end{minipage}
% \caption{Quantitative results on the dataset~\cite{Bagdanov2012Florence}. The first column presents an input image and its ground truth shape. In the other columns we show different face reconstructions with 2D: 2D fitting method, 2D+3D: joint 2D and 3D method, 2D+3D+W: 2D and 3D method with adaptive weights, 2D+3D+P+W: 2D and 3D method with pose refinement and adaptive weights. We also show  their corresponding error maps. In the bottom, we show the reconstruction error values.}
% \label{fig:OurCompare}
% \end{figure*}
\begin{table}[t]
\begin{center}
\caption{Quantitative evaluation on MICC dataset~\cite{Bagdanov2012Florence} with left and right non-frontal views. 2D: 2D fitting method; 2D+3D: joint 2D and 3D fitting method; 2D+3D+W: joint 2D and 3D fitting method with adaptive re-weighting; 2D+3D+P+W: joint 2D and 3D fitting method with pose refinement and adaptive re-weighting. } \label{tab:Our}
\normalsize
\renewcommand{\arraystretch}{1.5}
\begin{tabular}{|c|c|c|}
  \hline
  % after \\: \hline or \cline{col1-col2} \cline{col3-col4} ...
  Method & Left & Right
  \\
  \hline
  2D & 3.184 & 3.146 \\
  3D & 2.053 & 2.026 \\
  2D+3D & 2.002 & 2.241 \\
  2D+3D+W & 1.904 & 1.911 \\
  2D+3D+P+W & \textbf{1.812} & \textbf{1.835} \\
  \hline
\end{tabular}
\end{center}
\end{table}

\subsection{Results}
\label{sec:results}
We demonstrate the robustness and compatibility of our approach in Figure \ref{fig:result}. In order to prove our robustness to extreme poses and partial occlusions, we show more face reconstruction results on AFLW2000~\cite{Zhu2016Face} dataset in Figure~\ref{fig:bc3}. The first row shows the results of images with hair occlusion, the second and third rows are the results of images with small pose rotation, and the rest rows show the face reconstruction results with large poses and occlusion with glasses. Our method can accurately reconstruct the 3D faces for these challenging cases.

\begin{figure*}[!t]
	\centering
	\includegraphics[width=0.95\linewidth]{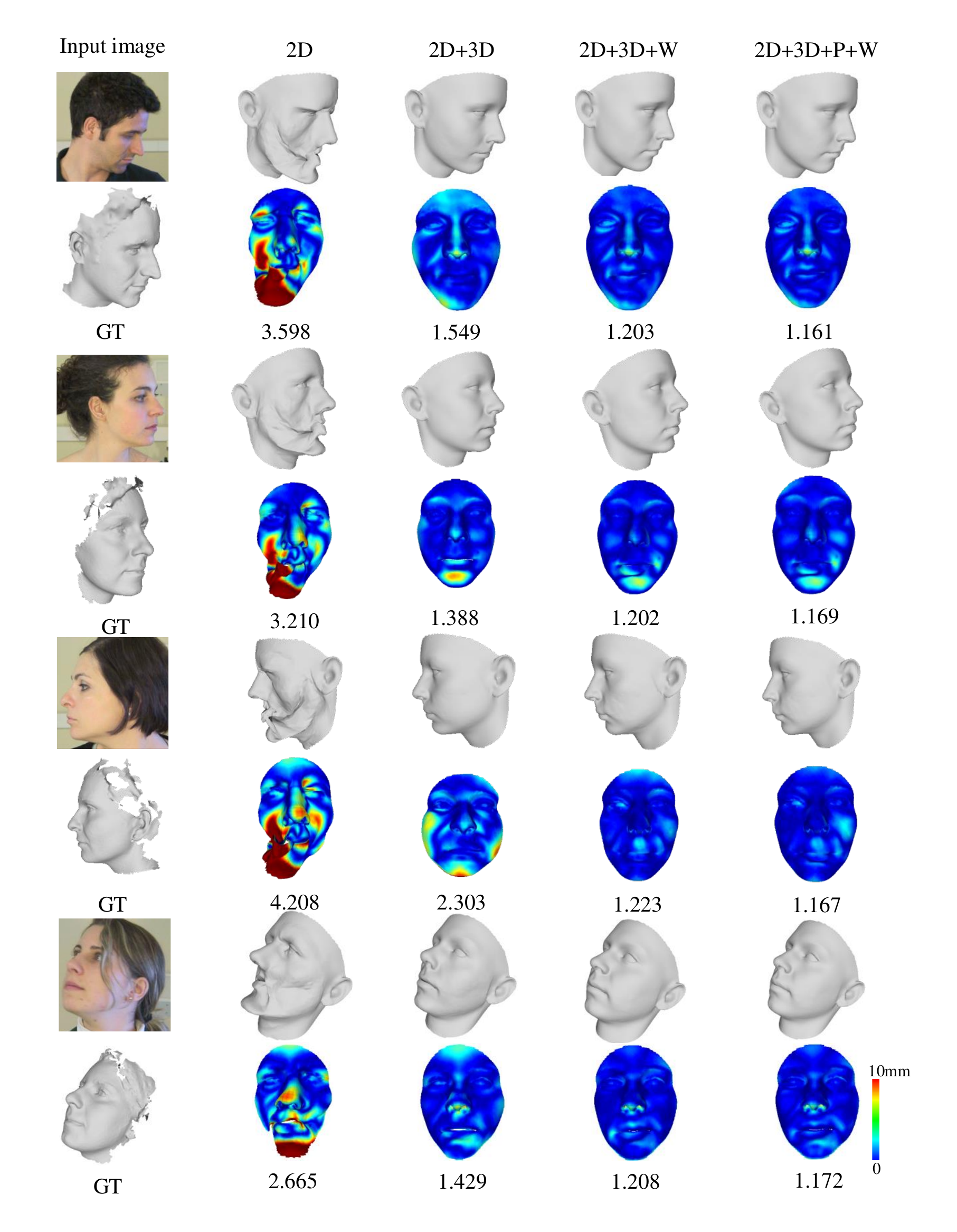}
	\caption{Qualitative results on the MICC dataset~\cite{Bagdanov2012Florence}. The first column presents input images together with the ground truths. The other columns show the face reconstruction results using 2D: 2D fitting method, 2D+3D: joint 2D and 3D fitting method, 2D+3D+W: joint 2D and 3D fitting method with adaptive re-weighting, and 2D+3D+P+W: joint 2D and 3D fitting method with pose refinement and adaptive re-weighting. }
	\label{fig:OurCompare}
\end{figure*}

\begin{figure*}[!t]
	\centering
	\includegraphics[width=0.6\linewidth]{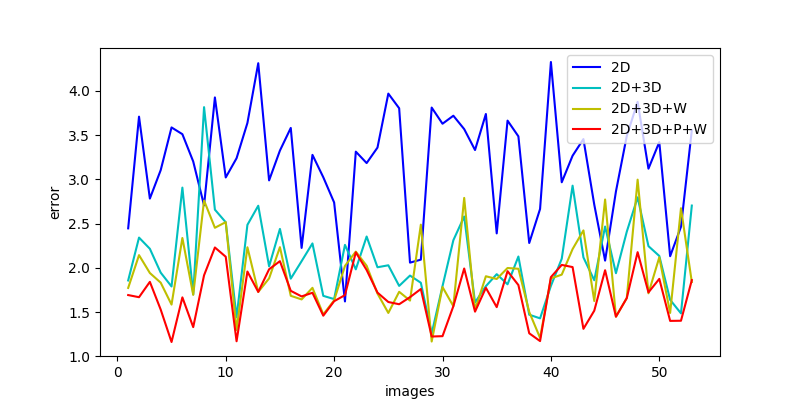}
	\caption{The reconstruction errors for left-view images. }
%%%YKL Please update the figure where x-axis should be images (NOT method!)
%%%YKL legend: 2, 23, 23w, 23pw should be changed to 2D, 2D+3D, 2D+3D+W, 2D+3D+P+W
%%%YJ modified
	\label{fig:lineour}
\end{figure*}

\begin{figure*}[!t]
  \centering
  \includegraphics[width=1.0\linewidth]{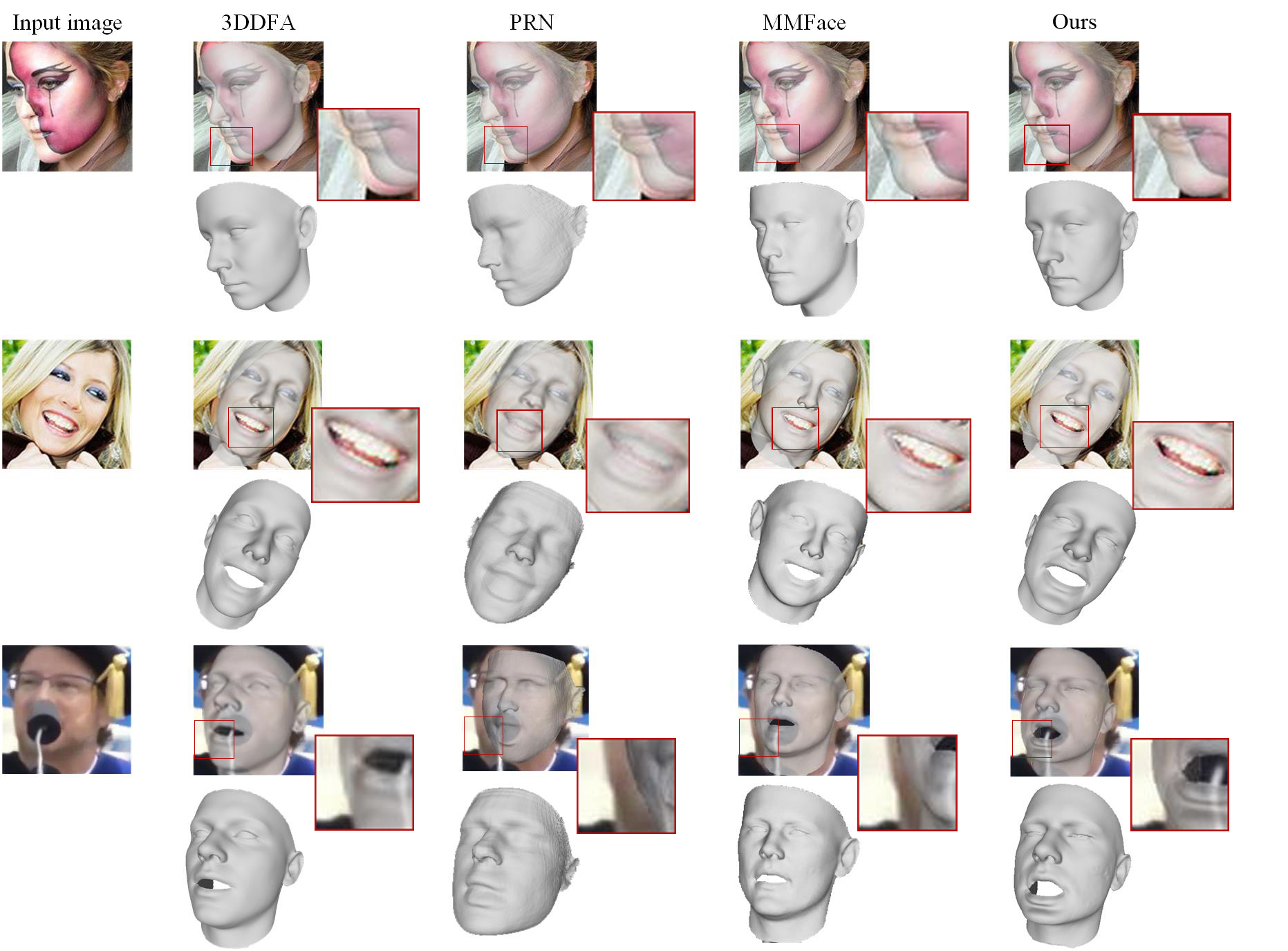}
  \caption{3D face reconstruction results compared with 3DDFA~\cite{Zhu2016Face}, PRN~\cite{Yao2018Joint} and MMFace~\cite{yi2019mmface}. The first two images are from AFLW2000~\cite{Zhu2016Face} and the last image comes from 300VW-3D~\cite{Bulat2017How}. }
  \label{fig:OtherCompare}
\end{figure*}

\subsection{Ablation Study}\label{sec:abl_study}

To prove the performance improvement of our method for partial occlusions and extreme poses, we select 106 non-frontal images with left or right view direction (53 images per case) for each subject from the MICC dataset~\cite{Bagdanov2012Florence} as our test dataset in this section. Table~\ref{tab:Our} shows the quantitative results compared with four variants. Joint 2D and 3D fitting is better 2D fitting for the left case, but is slightly worse than 3D fitting for the right case due to some improper weighting. The adaptive re-weighting improves the performance by $15\%$ for the right case. Higher accuracy can be achieved by adding our coarse-to-fine pose estimation.
We also show the detailed error results for each left-view image in Figure~\ref{fig:lineour}.

The visual results of different variants are shown in Figure~\ref{fig:OurCompare}. The reconstruction errors are color-coded on the reconstructed model for visual inspection, and the average error is given below each case. It can be seen that it is difficult to recover an accurate model with only 2D landmarks due to lack of the depth information. Combining 3D landmarks with the corresponding 2D landmarks can largely improve the reconstruction accuracy. Adaptive re-weighting further improves the performance. The most satisfactory result is achieved by adding the coarse-to-fine pose estimation method.

\subsection{Comparison}
\label{sec:compare}

\subsubsection{Qualitative Evaluation}\label{qual_evaluation}

In order to prove the reliability of our method, we compare our method with three state-of-the-art face reconstruction approaches, 3DDFA~\cite{Zhu2016Face}, PRN~\cite{Yao2018Joint} and MMFace~\cite{yi2019mmface}, on the AFLW2000~\cite{Zhu2016Face}, 300VW-3D~\cite{Bulat2017How} and MICC~\cite{Bagdanov2012Florence} datasets. Because MMface~\cite{yi2019mmface} does not publish the code, we only make a qualitative comparison with this method by using the images provided in their paper. Figure~\ref{fig:OtherCompare} illustrates the qualitative evaluation results compared with these methods. In order to show the reconstruction results consistent with MMface~\cite{yi2019mmface}, we reduce the transparency of the model to overlay on the image, which can better observe the correctness of the eyes, mouth, nose, and face shape of the model compared with the input image. It can be seen that 3DDFA~\cite{Zhu2016Face} can reconstruct fine models but the results are all similar to the mean face and lack the consistency with the image. PRN~\cite{Yao2018Joint} can estimate accurate face orientation but fails to reconstruct fine facial geometry. MMFace~\cite{yi2019mmface} cannot recover the contour well, \eg, the third image. Our method estimates better poses and reconstructs more accurate face models, benefitting from our coarse-to-fine pose estimation and joint 2D and 3D optimization with adaptive re-weighting.

\begin{table}[t]
\begin{center}
\caption{Comparison of 3D face reconstruction on MICC dataset~\cite{Bagdanov2012Florence}.} \label{tab:Other}
\normalsize
\renewcommand{\arraystretch}{1.5}
\begin{tabular}{|c|c|c|}
  \hline
  % after \\: \hline or \cline{col1-col2} \cline{col3-col4} ...
  Method & Frontal & Non-frontal
  \\
  \hline
  3DDFA~\cite{Zhu2016Face} & 2.244 & 2.379 \\
  PRN~\cite{Yao2018Joint} & 2.086 & 1.934 \\
  Ours & \textbf{1.819} & \textbf{1.770} \\
  \hline
\end{tabular}
\end{center}
\end{table}

\begin{figure*}[htbp]
  \centering
  \includegraphics[width=0.9\linewidth]{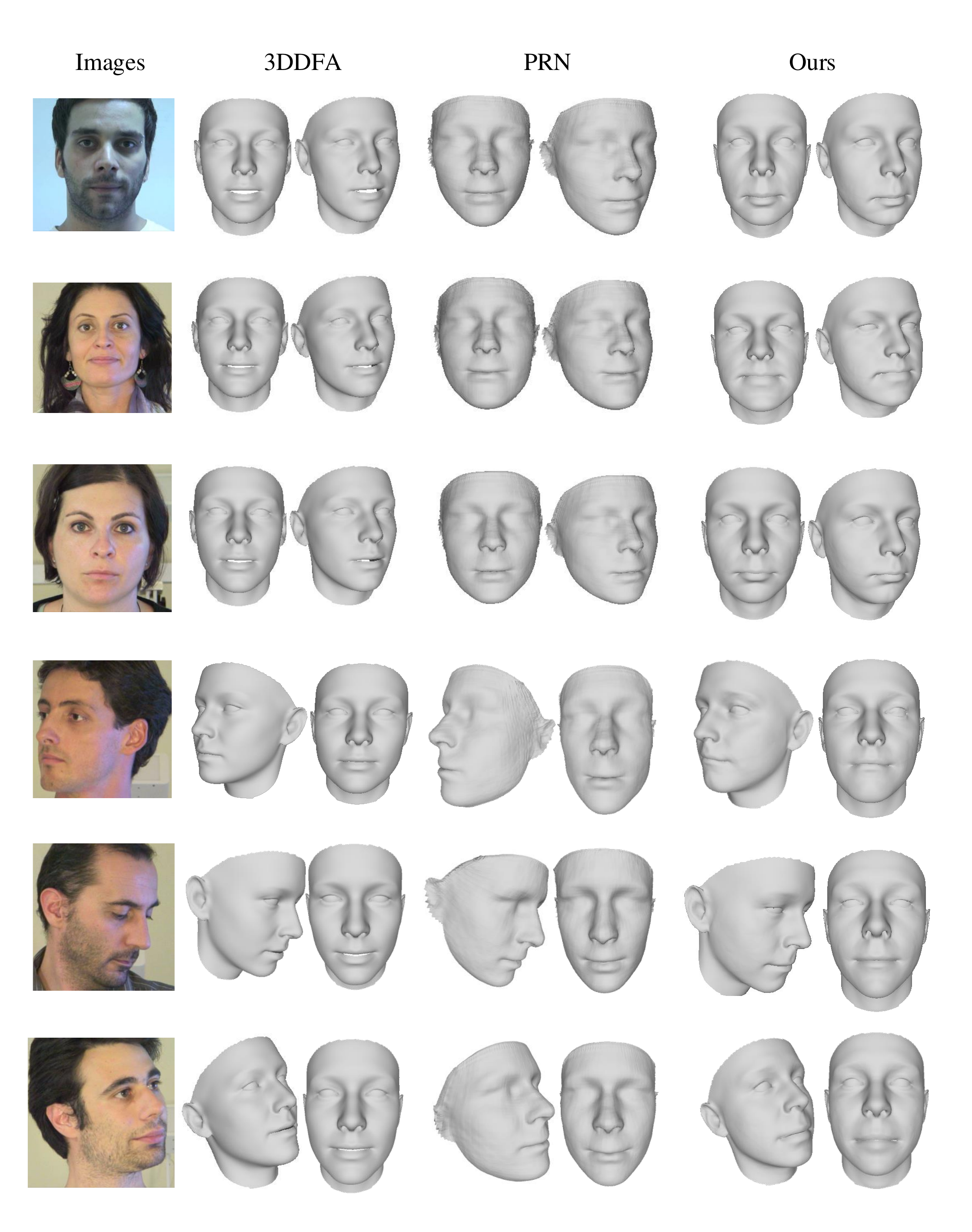}
  \caption{More 3D face reconstruction results compared with 3DDFA~\cite{Zhu2016Face} and PRN~\cite{Yao2018Joint}.}
  \label{fig:bc1}
\end{figure*}

\begin{figure*}[ht]
	\centering
	\includegraphics[width=0.7\linewidth]{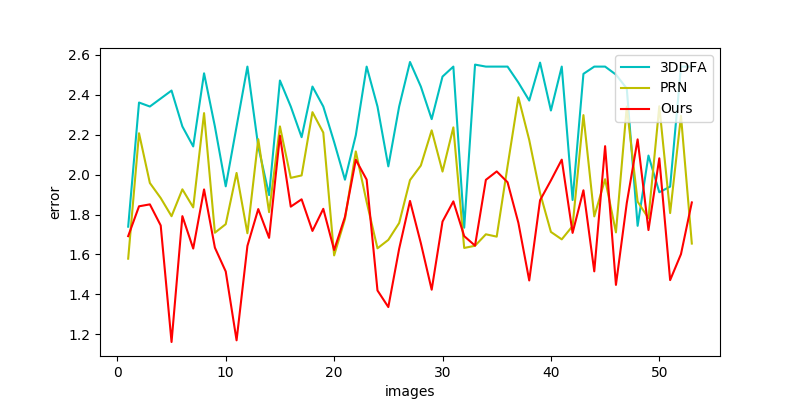}
	\caption{Quantitative results on the non-frontal images.}
	\label{fig:othernonline}
\end{figure*}

\begin{figure*}[ht]
	\centering
	\includegraphics[width=0.6\linewidth]{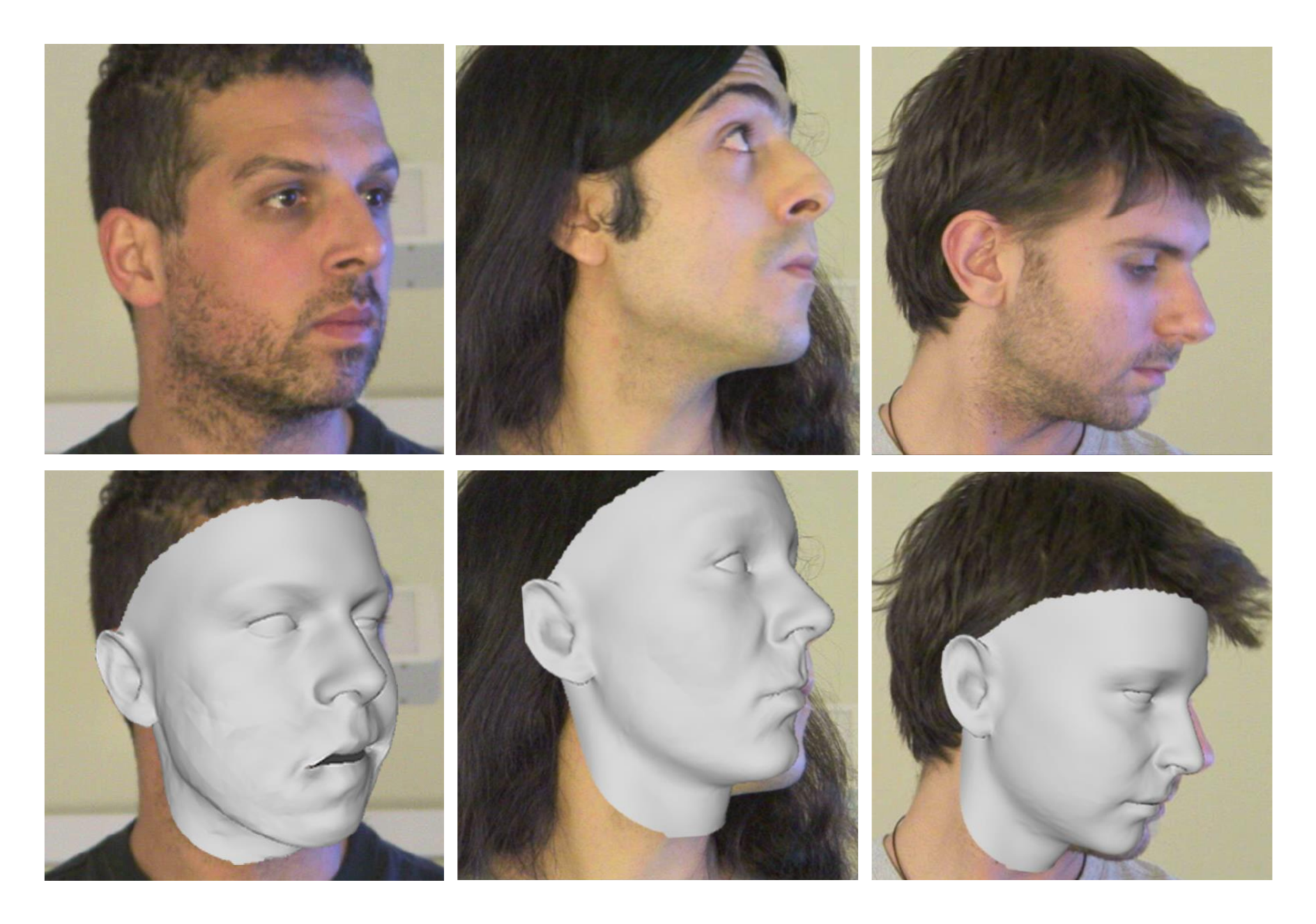}
	\caption{Examples of failure cases.}
	\label{fig:failure}
\end{figure*}

We further compare our method with 3DDFA~\cite{Zhu2016Face} and PRN~\cite{Yao2018Joint} on the MICC dataset~\cite{Bagdanov2012Florence} in Figure~\ref{fig:bc1}. MMface~\cite{yi2019mmface} does not publish the code and hence we cannot test more images with this method. For each reconstructed model, we show a frontal view and a non-frontal view for more intuitive observation. It can be observed that 3DDFA~\cite{Zhu2016Face} can reconstruct a fine model but with inaccurate mouth estimation. PRN~\cite{Yao2018Joint} recovers unreasonable face shapes, especially for the second and fourth images. On the contrary, our method reconstructs more accurate face models for both frontal and non-frontal images.

\subsubsection{Quantitative Evaluation}\label{quan_evaluation}
To evaluate the performance on different cases, we choose 53 frontal images and 53 non-frontal images for each subject from the MICC dataset~\cite{Bagdanov2012Florence} as our test datasets. Table~\ref{tab:Other} shows quantitative results compared with 3DDFA~\cite{Zhu2016Face} and PRN~\cite{Yao2018Joint} on the test datasets. Because MMface~\cite{yi2019mmface} does not publish the code, we cannot compare this method quantitatively. As shown in this table, our method achieves smaller average errors for both frontal and non-frontal datasets than the other methods, especially for the occluded images. The detailed errors for each non-frontal image are shown in Figure~\ref{fig:othernonline}. It can be seen that our method consistently has less error for face reconstruction.

% if have a single appendix:
%\appendix[Proof of the Zonklar Equations]
% or
%\appendix  % for no appendix heading
% do not use \section anymore after \appendix, only \section*
% is possibly needed

% use appendices with more than one appendix
% then use \section to start each appendix
% you must declare a \section before using any
% \subsection or using \label (\appendices by itself
% starts a section numbered zero.)
%

\section{Conclusion and Discussion}
\label{sec:conclude}
In this paper, we propose a novel method to solve the challenges of face reconstruction from a single image under partial occlusions and large poses. First, we propose a coarse-to-fine pose estimation method, which divides pose estimation into two steps to improve the accuracy. Second, we propose a novel joint 2D and 3D optimization method with adaptive re-weighting. Our pose estimation and the 2D and 3D weight adaptation are considered as a bundle, and solved by a joint optimization algorithm. Experimental results on public datasets demonstrate that our method can reconstruct more accurate face geometry consistent with the images even for occlusions or extreme poses, compared with the state-of-the-art methods.

Figure~\ref{fig:failure} shows some failure examples using our method due to wrong estimation of 2D and 3D landmarks for occlusion cases. In the future work, we will try to improve the accuracy of landmark detection with the help of face reconstruction iteratively. Also, we will try to combine our optimization method with learning-based prior or representation.

\bibliographystyle{elsarticle-num}
% argument is your BibTeX string definitions and bibliography database(s)
\bibliography{yjref}

% that's all folks
\end{document}